\documentclass[10pt,twocolumn,letterpaper]{article}

\usepackage[pagenumbers]{wacv} 

\usepackage{appendix}
\usepackage{booktabs}
\usepackage{epsfig}
\usepackage{graphicx}
\usepackage{amsmath}
\usepackage{amssymb}
\usepackage{MnSymbol}
\usepackage{multirow}
\usepackage{pifont}
\usepackage{multicol}
\newcommand{\cmark}{\ding{51}}%
\newcommand{\xmark}{\ding{55}}%

\usepackage{tabulary}
\usepackage{color, colortbl}
\definecolor{Gray}{gray}{0.95}
\usepackage[dvipsnames]{xcolor}

\usepackage{algorithm}
\usepackage{algpseudocode}

\usepackage[pagebackref,breaklinks,colorlinks]{hyperref}

\usepackage[capitalize]{cleveref}
\crefname{section}{Sec.}{Secs.}
\Crefname{section}{Section}{Sections}
\Crefname{table}{Table}{Tables}
\crefname{table}{Tab.}{Tabs.}

\DeclareMathOperator*{\argmin}{arg\,min}

\def\wacvPaperID{1501} 
\def\confName{WACV}
\def\confYear{2024}

\begin{document}

\title{A Coarse-to-Fine Pseudo-Labeling (C2FPL) Framework for Unsupervised Video Anomaly
Detection}

\author{Anas Al-lahham \quad Nurbek Tastan \quad Zaigham Zaheer \quad Karthik Nandakumar \\ 
Mohamed bin Zayed University of Artificial Intelligence (MBZUAI)\\ 
Abu Dhabi, UAE\\ 
{\tt\small \{anas.al-lahham, nurbek.tastan, zaigham.zaheer, karthik.nandakumar\}@mbzuai.ac.ae}} 
\maketitle

\begin{abstract}
Detection of anomalous events in videos is an important problem in applications such as surveillance. Video anomaly detection (VAD) is well-studied in the one-class classification (OCC) and weakly supervised (WS) settings. However, fully unsupervised (US) video anomaly detection methods, which learn a complete system without any annotation or human supervision, have not been explored in depth. This is because the lack of any ground truth annotations significantly increases the magnitude of the VAD challenge. To address this challenge, we propose a simple-but-effective two-stage pseudo-label generation framework that produces segment-level (normal/anomaly) pseudo-labels, which can be further used to train a segment-level anomaly detector in a supervised manner. The proposed coarse-to-fine pseudo-label (C2FPL) generator employs carefully-designed hierarchical divisive clustering and statistical hypothesis testing to identify anomalous video segments from a set of completely unlabeled videos. The trained anomaly detector can be directly applied on segments of an unseen test video to obtain segment-level, and subsequently, frame-level anomaly predictions. Extensive studies on two large-scale public-domain datasets, UCF-Crime and XD-Violence, demonstrate that the proposed unsupervised approach achieves superior performance compared to all existing OCC and US methods, while yielding comparable performance to the state-of-the-art WS methods. Code is available at: \href{https://github.com/AnasEmad11/C2FPL}{https://github.com/AnasEmad11/C2FPL} 
\end{abstract}

\section{Introduction}
\label{sec:intro}

\begin{figure}[t]
\begin{center}
    \includegraphics[width=1\linewidth]{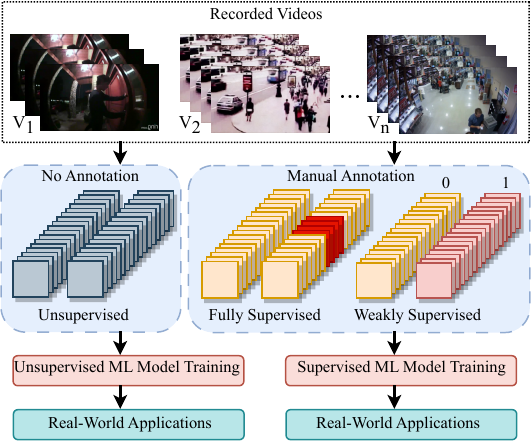}
\end{center}
   \caption{
Supervised (right) vs. unsupervised (left) video anomaly detection pipeline. In a supervised setting, some sort of manual annotation of the recorded videos is required for training an anomaly detection model. We aim to eliminate this annotation step by proposing a fully unsupervised approach.}
\label{fig:highlight}
\end{figure}

Applications such as video surveillance continuously generate large amounts of video data. While a vast majority of these videos only contain normal behavior, it is essential to detect anomalous events (e.g., shooting, road accidents, fighting, etc.) that deviate from normal behavior and may occur occasionally in such videos. Hence, video anomaly detection (VAD) is a critical problem, especially in surveillance applications \cite{mohammadi2016angry, sultani2018real, kamijo2000traffic,luo2017revisit}. 

Conventional VAD methods rely heavily on manually annotated anomaly examples (Figure \ref{fig:highlight}(right)) \cite{antic2011video}. However, given the rare occurrence and short temporal nature of anomalies in real-world scenarios, obtaining accurate fine-grained annotations is a laborious task. 
Recently, several VAD methods have been proposed to leverage video-level labels and perform weakly supervised (WS) training \cite{feng2021mist, morais2019learning, sultani2018real,zhong2019graph, zaheer2020claws,tian2021weakly} to reduce the annotation costs. However, since surveillance datasets are usually a large-scale collection of videos, it is still cumbersome to obtain any kind of labels. For example, to obtain even a video-level binary label, an annotator may still have to watch the whole video, which can take a considerable amount of time. For example, a well-known WS-VAD dataset called XD-Violence \cite{wu2020not} contains videos spanning $217$ hours. An alternative paradigm for VAD is one-class classification (OCC), which assumes that only normal videos are available for training \cite{sultani2018real,tian2021weakly,wu2022self,Purwanto2021ICCV,zaheer2020claws}. However, the OCC setting does not completely alleviate the annotation problem because an annotator still has to watch all the training videos to ensure that no anomaly is present within them.



A label-free fully unsupervised approach is a more practical and useful setting, especially in real-world scenarios where recording video data is easier than annotating it \cite{zaheer2022generative}.
An unsupervised video anomaly detection (US-VAD) method can address the aforementioned disadvantages of supervised methods by completely eradicating the need for manual annotations (Figure \ref{fig:highlight}). However, US-VAD methods are yet to gain much traction within the computer vision community. Recently, Zaheer \etal \cite{zaheer2022generative} introduced an US-VAD approach in which the model is trained on unlabeled normal and anomalous videos. Their idea is to utilize several properties of the training data to obtain pseudo-labels via cooperation between a generator and a classifier. While this method is elegant, its performance is significantly lower than the state-of-the-art WS and OCC methods \cite{thakare2023rareanom,zaheer2022generative}.

In this work, we attempt to bridge this gap between unsupervised and supervised methods by taking unlabelled set of training videos as input and producing segment-level pseudo-labels without relying on any human supervision. Towards this end, we make the following key contributions:

\begin{itemize}
    \item We propose a two-stage coarse-to-fine pseudo-label (C2FPL) generator that utilizes hierarchical divisive (top-down) clustering and statistical hypothesis testing to obtain segment-level (fine-grained) pseudo-labels.

    \item Based on the C2FPL framework, we propose an US-VAD system that is trainable without any annotations. To the best of our knowledge, this is among the first few works to explore the US-VAD setting in detail.
    
    \item We evaluate the proposed approach on two large-scale VAD datasets, UCF-Crime \cite{sultani2018real} and XD-Violence \cite{wu2020not}, and achieve state-of-the-art performance in the unsupervised category, while also outperforming all existing OCC and several WS-VAD methods.
\end{itemize}

\begin{figure*}[t]
\begin{center}
    \includegraphics[width=1\linewidth]{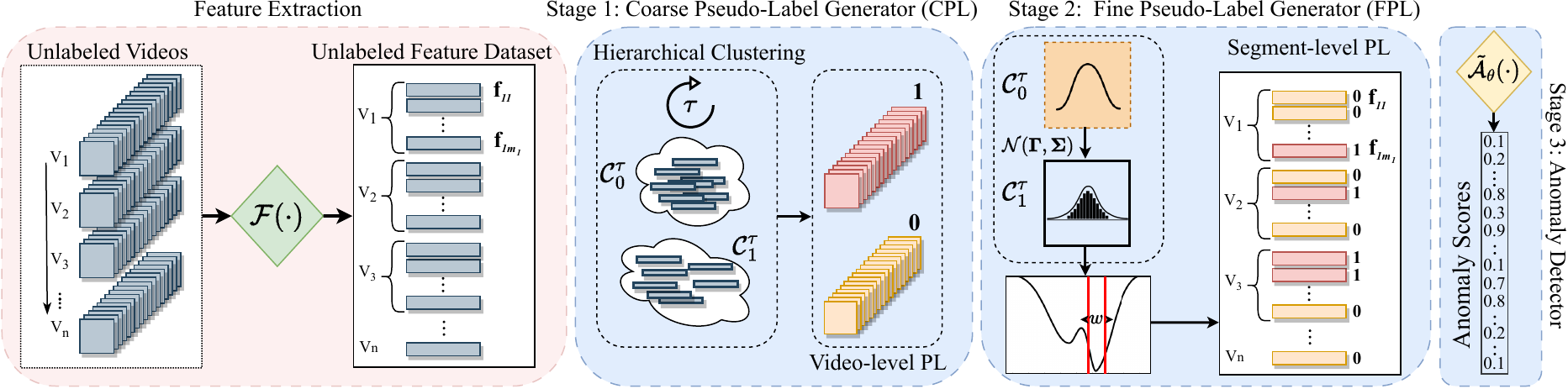}
\end{center}
   \caption{Overall training pipeline of the proposed C2FPL framework for unsupervised video anomaly detection (US-VAD). All training videos are first partitioned into segments and each segment is represented by a feature vector obtained using a pre-trained feature extractor. Then, our two-stage coarse-to-fine pseudo-label (C2FPL) generator produces segment-level pseudo-labels, which are used to train a segment-level anomaly detector. Pseudo-label generation consists of two stages: hierarchical divisive clustering that generates coarse (video-level) pseudo-labels (CPL) and statistical hypothesis testing that creates fine-grained (segment-level) pseudo-labels (FPL).}
\label{fig:arch}
\end{figure*}

\section{Related Work}
Early VAD methods mostly relied on supervised learning, where anomalous frames in a video are explicitly labeled in the training data \cite{tschuchnig2022anomaly,gornitz2013toward}. Since supervised approaches require large amounts of annotated data and annotation of anomalies is a laborious task, WS, OCC, and US VAD methods are gaining more attention.

\subsection{One-Class Classification for VAD}

To avoid the capturing of anomalous examples, researchers have widely explored one-class classification (OCC) methods \cite{zaheer2020old,hasan2016anomaly,lu2013abnormal,wang2019gods}. In OCC-VAD, only normal videos are used to train an outlier detector. At the time of inference, data instances that do not conform to the learned normal representations are predicted as anomalous. Since OCC methods are known to fail if normal data contains some anomaly examples \cite{zaheer2022generative}, they require careful verification of all the videos in the dataset, which does not reduce the annotation load. Furthermore, video data is often too diverse to be modeled successfully and new normal scenes differing from the learned representations may be classified as anomalous. Therefore, OCC approach has limited applicability in the context of VAD.

\subsection{Weakly Supervised VAD}
Taking advantage of weakly labeled (i.e., video-level labels) anomalous samples has led to significant improvements over OCC training \cite{tian2021weakly, sultani2018real}. Multiple Instance Learning (MIL) is one of the most commonly used methods for WS-VAD \cite{sultani2018real, tian2021weakly, wu2022self, Purwanto2021ICCV}, where segments of a video are grouped into a bag and bag-level labels are assigned. Sultani \etal. \cite{sultani2018real} first introduced the MIL framework with a ranking loss function, which is computed between the top-scoring segments of normal and anomaly bags. 

One of the key challenges in WS-VAD is that the positive (anomaly) bags are noisy. Since anomalies are localized temporally, most of the segments in an anomaly bag are also normal. Therefore, Zhong \etal. \cite{zhong2019graph} reformulated the problem as binary classification in the presence of noisy labels and used a graph convolution network (GCN) to remove label noise. The training of GCN was computationally expensive due to the presence of an action classifier. 
Furthermore, MIL-based methods require complete video inputs at each training iteration. Consequently, the correlation of the input data significantly affects the training of an anomaly detection network. To minimize this correlation, CLAWS Net \cite{zaheer2020claws} proposed a random batch selection approach in which temporally consistent batches are arbitrarily selected for training a binary classifier.


\subsection{Unsupervised VAD} 

Unsupervised video anomaly detection (US-VAD) methods are learned using unlabeled training data. This problem is extremely challenging due to the lack of ground truth supervision and the rarity of anomalies. However, it is highly rewarding because it can completely eradicate the costs associated with obtaining manual annotations and allow such systems to be deployed without human intervention. Due to the difficulty of the problem, it has received little attention in the literature. Generative Cooperative Learning \cite{zaheer2022generative} is a recent work that presents an US-VAD system to detect anomalies in a given video by first training a generative model to reconstruct normal video frames and then using the discrepancy between the reconstructed frames and the actual frames as a measure of anomaly. It involves training two models simultaneously: one to reconstruct the normal frames and the other to generate classification scores. 

\section{Proposed Methodology}

\noindent \textbf{Problem Definition}: Let $\mathcal{D} = \{V_1, V_2, \cdots, V_n\}$ be a training dataset containing $n$ videos without any labels. The goal of US-VAD is to use $\mathcal{D}$ and learn an anomaly detector $\mathcal{A}(\cdot)$ that classifies each frame in a given test video $V_*$ as either \textit{normal} ($0$) or \textit{anomalous} ($1$). 

\noindent \textbf{Notations}: We split each video $V_{i}$ into a sequence of $m_{i}$ non-overlapping segments $S_{ij}$, where each segment is in turn composed of $r$ frames. Note that $i \in [1,n]$ refers to the video index, and $j \in [1,m_{i}]$ is the segment index within a video. While many WS-VAD methods \cite{sultani2018real,tian2021weakly,wu2022self, Purwanto2021ICCV} compress each video into a fixed number of segments (i.e., $m_i =m, \forall i \in [1,n]$) along the temporal axis, we avoid any compression and make use of all available non-overlapping segments. For each segment $S_{ij}$, a feature vector $\mathbf{f}_{ij} \in \mathbb{R}^{d}$ is obtained using a pre-trained feature extractor $\mathcal{F}(\cdot)$.

\noindent \textbf{High-level Overview of the Proposed Solution}: Our coarse-to-fine pseudo-labeling (C2FPL) framework for US-VAD consists of three main stages during training (see Figure \ref{fig:arch}). In the first coarse pseudo-labeling (CPL) stage, we generate a video-level pseudo-label $\hat{y}_i \in \{0,1\}$, $i \in [1,n]$ for each video in the training set using a hierarchical divisive clustering approach. In the second fine pseudo-labeling (FPL) stage, we generate segment-level pseudo-labels $\tilde{y}_{ij} \in \{0,1\}$, $i \in [1,n]$, $j \in [1,m_{i}]$ for all the segments in the training set through statistical hypothesis testing. In the third anomaly detection (AD) stage, we train a segment-level anomaly detector $\tilde{\mathcal{A}}_{\theta}(\cdot): \mathbb{R}^{d} \rightarrow [0,1]$ that assigns an anomaly score between $0$ and $1$ (higher values indicate higher confidence of being an anomaly) to the given video segment based on its feature representation $\mathbf{f}_{ij}$.

\subsection{Coarse (Video-Level) Pseudo-Label Generator}

Since the training dataset does not contain any labels, we first generate pseudo-labels for the videos in the training set by recursively clustering them into two groups: normal and anomalous (see Alg. \ref{alg:c2fpl}). The idea of using iterative clustering to generate pseudo-labels has been considered earlier in other application domains \cite{zhang2021refining,cho2022part,ahmed2021adaptive }. However, direct application of these methods to the US-VAD problem fails to provide satisfactory solutions due to two reasons. Firstly, directly clustering multivariate features $\mathbf{f}_{ij}$ leads to a curse of dimensionality (features are high-dimensional but the sample size is small). Secondly, the clusters in our context are not permutation-invariant (normal and anomalous cluster labels cannot be interchanged). To overcome these problems, we propose a method that relies on a low-dimensional feature summary and divisive hierarchical clustering. 

Previous works in WS-VAD have shown that normal video segments have lower temporal feature magnitude compared to anomalous segments \cite{tian2021featuremagnitude}. Furthermore, we also observed that the variations in feature magnitude across different segments are lower for normal videos. Based on this intuition, we represent each video $V_i$ using a statistical summary $\mathbf{x}_i = [\mu_i, \sigma_i]$ of its features as follows:

\begin{equation}
    \mu_i = \frac{1}{m_i} \sum_{j=1}^{m_i}||\mathbf{f}_{ij}||_2, 
    \label{eq:mu}
\end{equation}

\begin{equation}
    \sigma_i = \sqrt{\frac{1}{(m_i-1)} \sum_{j=1}^{m_i}(||\mathbf{f}_{ij}||_2 - \mu_i)^2}, 
    \label{eq:sigma}
\end{equation}

\noindent where $||\cdot||_2$ represents the $\ell_2$ norm of a vector. Thus, each video $V_i$ is represented using a 2D vector $\mathbf{x}_i$, corresponding to the mean and standard deviation of the feature magnitude of its segments. This ensures a uniform representation of all videos despite their varying temporal length.

Videos in the training set are iteratively divided into two clusters ($\mathcal{C}_0^t$ and $\mathcal{C}_1^t$) based on the above representation $\mathbf{x}_i$. Here, $t$ denotes the step index and $\mathcal{C}_0$ and $\mathcal{C}_1$ represent the normal and anomaly clusters, respectively. Since no data labels are available, assigning normal and anomaly labels to the clusters is not trivial. Intuitively, easy anomalies (considered as easy outliers) may be separated into a smaller cluster. On the other hand, the larger cluster is likely to contain more normal videos as well as some hard anomalies that need further refinement. Therefore, initially, all the videos in the training set are assigned to the normal cluster and the anomaly cluster is initialized to an empty set, i.e., $\mathcal{C}_0^0 = \{\mathbf{x}_i\}_{i \in [1,n]}$ and $\mathcal{C}_1^0 = \emptyset$. At each step $t$ ($t \geq 1$), the cluster $\mathcal{C}_0^{t-1}$ is re-clustered to obtain two new child clusters, say $\mathcal{C}_l$ and $\mathcal{C}_s$ with $|\mathcal{C}_l|$ and $|\mathcal{C}_s|$ samples, respectively. Without loss of generality, let $|\mathcal{C}_s| < |\mathcal{C}_l|$. The smaller cluster $\mathcal{C}_s$ is merged with the previous anomaly cluster, i.e., $\mathcal{C}_1^{t} = (\mathcal{C}_1^{t-1}~\cup~\mathcal{C}_s)$, while the larger cluster is labeled as normal, i.e., $\mathcal{C}_0^{t} = \mathcal{C}_l$. This process is repeated until the ratio of the number of videos in the anomaly cluster ($|\mathcal{C}_1^t|$) to the number of videos in the normal cluster ($|\mathcal{C}_0^t|$) is larger than a threshold, i.e., $\frac{|\mathcal{C}_1^t|}{|\mathcal{C}_0^t|} > \eta$. At the end of the CPL stage, all the videos in the training set are assigned a pseudo-label based on their corresponding cluster index, i.e., $\hat{y}_i = k$, if $\mathbf{x}_i \in \mathcal{C}_k^{\tau}$, where $k \in \{0,1\}$ and $\tau$ denotes the final clustering iteration.

\begin{algorithm}
\caption{Coarse-to-Fine Pseudo-Label Generation}
\label{alg:c2fpl}
\begin{algorithmic}[1]
\Statex \textbf{Input:} Training dataset $\mathcal{D} = \{V_1, \cdots , V_n\}$, pre-trained feature extractor $\mathcal{F}(\cdot)$, parameters $\eta$, $\beta$
\Statex \textbf{Output:} Segment-level pseudo-labels $\{ \tilde{y}_{ij}\}, \text{where } i \in [1,n] \text{ and } j \in [1,m_i]$ 



\For{$i = 1$ to $n$} 
    \State Partition $V_i$ into $m_i$ segments $[S_{i1},\cdots,S_{im_i}]$
    \State Extract segment features $[\mathbf{f}_{i1},\cdots, \mathbf{f}_{im_i}]$ using $\mathcal{F}(\cdot)$
    \State Compute $\mathbf{x}_i = [\mu_i, \sigma_i]$ using Eqs. \ref{eq:mu} \& \ref{eq:sigma}
\EndFor 

\State \textcolor{blue}{\textbf{CPL}}: $t=0$, $\mathcal{C}_0^t = \{\mathbf{x}_1, \cdots , \mathbf{x}_n\}$, $\mathcal{C}_1^t = \emptyset$
\While{$|\mathcal{C}_1^t|$ / $|\mathcal{C}_0^t| \leq \eta$} 
    \State $(\mathcal{C}_s, \mathcal{C}_l) \gets$ Clustering($\mathcal{C}_0^t$), where $|\mathcal{C}_s| < |\mathcal{C}_l|$

    \State $\mathcal{C}_1^{t+1} \gets \mathcal{C}_1^t \cup \mathcal{C}_s$, $\mathcal{C}_0^t \gets \mathcal{C}_l$ 
    \State $t \gets t + 1$ 
\EndWhile 
\State $\forall i \in [1,n], \hat{y}_i \gets 0$ \textbf{if} {$\mathbf{x}_i \in \mathcal{C}_0^{t}$}, \textbf{else} $\hat{y}_i \gets 1$
\State \textcolor{blue}{\textbf{FPL}}: $\forall i \in [1,n], j \in [1,m_i]$, $\tilde{y}_{ij} \gets 0$, Compute $\mathbf{z}_{ij}$
\State Compute $(\mathbf{\Gamma}, \mathbf{\Sigma})$ using Eqs. \ref{eq:gamma} \& \ref{eq:Sigma}





\For{$i = 1$ to $n$} 
    \If{$\hat{y}_i = 1$}
        \State Compute $p_{ij}$ using Eq. \ref{eq: pdf}, $\forall j \in [1,m_i]$
        \State $w_i \gets \lceil\beta m_i\rceil$
        \State $l_i  = \argmin_{l} \left\{ \frac{1}{w_i} \sum_{j=(l+1)}^{(l+w_i)} p_{ij}, ~ \forall ~ l \in [0,m_i-w_i] \right\}$
        \State $\tilde{y}_{ij} \gets 1, ~\forall~j \in [l_i+1,l_i+w]$
    \EndIf
\EndFor


\end{algorithmic}
\end{algorithm}

\subsection{Fine (Segment-Level) Pseudo-Label Generator}
All the segments from videos that are ``pseudo-labeled'' as normal ($\hat{y}_i=0$) by the previous stage can be considered as normal. However, most of the segments in an anomalous video are also normal due to temporal localization of anomalies. Hence, further refinement of the coarse (video-level) labels is required to generate segment-level labels for anomalous videos. To achieve this goal, we treat the detection of anomalous segments as a statistical hypothesis testing problem. Specifically, the null hypothesis is that a given video segment is normal. By modeling the distribution of features under the null hypothesis as a Gaussian distribution, we identify the anomalous segments by estimating their p-value and rejecting the null hypothesis if the p-value is less than the significance level $\alpha$.

To model the distribution of features under the null hypothesis, we consider only the segments from videos that are pseudo-labeled as normal by the CPL stage. Let $\mathbf{z}_{ij} \in \mathbb{R}^{\tilde{d}}$ be a low-dimensional representation of a segment $S_{ij}$. We assume that $\mathbf{z}_{ij}$ follows a Gaussian distribution $\mathcal{N}(\mathbf{\Gamma},\mathbf{\Sigma})$ under the null hypothesis and estimate the parameters $\mathbf{\Gamma}$ and $\mathbf{\Sigma}$ as follows:

\begin{equation}
    \mathbf{\Gamma} = \frac{1}{M_0} \sum_{i=1, \hat{y}_i = 0}^{n}\sum_{j=1}^{m_i}\mathbf{z}_{ij}, 
    \label{eq:gamma}
\end{equation}

\begin{equation}
    \mathbf{\Sigma} = \frac{1}{(M_0-1)} \sum_{i=1, \hat{y}_i = 0}^{n}\sum_{j=1}^{m_i} (\mathbf{z}_{ij} - \mathbf{\Gamma})(\mathbf{z}_{ij} - \mathbf{\Gamma})^T, 
    \label{eq:Sigma}
\end{equation}

\noindent where $M_0 = \sum_{i=1, \hat{y}_i = 0}^n m_i$. Subsequently, for all the segments in videos that are pseudo-labeled as anomalous, the $p$-value is computed as:

\begin{equation}
    \label{eq: pdf}
    \begin{split}
          p_{ij} = \frac{1}{(2\pi)^{(\tilde{d}/2)} \sqrt{|\Sigma|}} 
          \exp \left(-\frac{1}{2} (\mathbf{z}_{ij} - \mathbf{\Gamma})^T\Sigma^{-1}(\mathbf{z}_{ij} - \mathbf{\Gamma}) \right),  
    \end{split}
\end{equation}

\begin{figure}[t]
    \centering
    \includegraphics[width=0.9\linewidth]{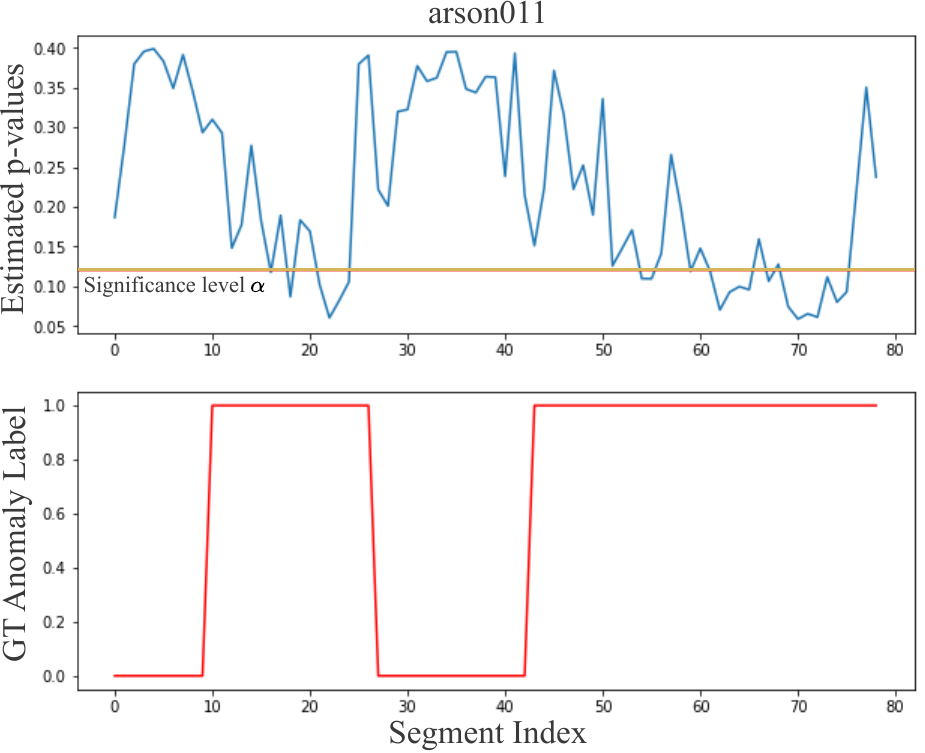}
    \caption{Statistical hypothesis testing approach applied to pseudo-label an anomalous video \textbf{arson011} from the \textit{validation set}. The top row shows the segment-level p-values for all the segments in the video, where a lower p-value means less likelihood of being normal. A possible pseudo-labeling strategy is to mark all segments with a p-value lower than the significance level (denoted by the horizontal orange line) as anomalous. Strong agreement can be observed between the estimated p-values and the ground-truth (GT) anomaly label for the given video shown in the bottom row.}
    \label{fig: pseudo-labeling}
\end{figure}

\noindent $\forall j \in [1,m_i], i \in [1,n]$ such that $\hat{y}_i = 1$. If $p_{ij} < \alpha$, the segment can be potentially assigned a pseudo-label of $1$. Figure \ref{fig: pseudo-labeling} shows an illustration of this approach, which clearly indicates strong agreement between the estimated p-values and the ground truth anomaly labels of the validation set.

One unresolved question in the above formulation is how to obtain the low-dimensional representation $\mathbf{z}_{ij}$ for a segment $S_{ij}$. In this work, we simply set $\mathbf{z}_{ij} = ||\mathbf{f}_{ij}||_2$ and hence $\tilde{d}=1$. Note that other statistics could also be employed in addition to (or in lieu of) the $\ell_2$ feature magnitude. 

Directly assigning a pseudo-label to a segment based on its p-value ignores the reality that anomalous segments in a video tend to be temporally contiguous. One way to overcome this limitation is to mark a contiguous sequence of $w_i = \lceil \beta m_i \rceil$ segments, $0 < \beta < 1$ and $\lceil \cdot \rceil$ represents the ceil function, as the anomalous region within each video that is pseudo-labeled as an anomaly. The anomalous region is determined by sliding a window of size $w_i$ across the video and selecting the window that has the lowest average p-values (i.e., $\min_{l} \left\{ \frac{1}{w_i} \sum_{j=(l+1)}^{(l+w_i)} p_{ij}, ~ \forall ~ l \in [0,m_i-w_i] \right\}$). Each segment present in this anomalous region is assigned a pseudo-label of $1$, while all the remaining segments are pseudo-labeled as normal (value of $0$). At the end of this FPL stage, a pseudo-label $\tilde{y}_{ij} \in \{0,1\}$ is assigned to all the segments in the training set.

\subsection{Anomaly Detector} \label{BBN}

The coarse and fine pseudo-label generators together provide a pseudo-label for every video segment in the training dataset. This results in a pseudo-labeled training set $\tilde{\mathcal{D}} = \{(\mathbf{f}_{ij},\tilde{y}_{ij})\}$ containing $M$ samples, where $i \in [1,n]$, $j \in [1,m_i]$, and $M = \sum_{i=1}^n m_i$. This labeled training set $\tilde{\mathcal{D}}$ can be used to train the anomaly detector $\tilde{\mathcal{A}}_{\theta}(\cdot)$ in a supervised fashion by minimizing the following objective:

\begin{equation}
    \min_{\theta} \sum_{i=1}^{n}\sum_{j=1}^{m_i} \mathcal{L}(\mathcal{A}_{\theta}(\mathbf{f}_{ij}),\tilde{y}_{ij}),
\end{equation}

\noindent where $\mathcal{L}$ is an appropriate loss function and $\theta$ denotes the parameters of the anomaly detector $\tilde{\mathcal{A}}(\cdot)$.

Following recent state-of-the-art methods \cite{sultani2018real,zaheer2022generative,tian2021weakly, zaheer2020claws}, 
two basic neural network architectures are considered for our anomaly detector. In particular, we employ a shallow neural network (Figure \ref{fig:arch}) with two fully connected (FC) hidden layers and one output layer mapped to a binary class. A dropout layer and a ReLU activation function are applied after each FC layer. Additionally, following Zaheer \etal \cite{zaheer2020claws}, we add two self-attention layers (detailed architecture is provided in the Supplementary material). A softmax activation function follows each of the self-attention layers, each of which has the same dimensions as the corresponding FC layer in the backbone network. Final anomaly score prediction is produced by the output sigmoid function. 

Unlike many existing methods (e.g., \cite{sultani2018real}) that require having a complete video in one training batch, our approach allows random segment selection for training. Recently, Zaheer \etal \cite{zaheer2020claws} demonstrated the benefits of feature vector randomization for training. However, based on its design, their method was limited to randomizing consecutive batches while maintaining the temporal order of segments within a batch. In our case, since we have obtained pseudo-labels for each segment, we can apply training with complete randomization to reap maximum benefits. Therefore, feature vectors are obtained across the dataset to form the training batches. Formally, each training batch $\mathcal{B}$ contains $B$ randomly selected samples from the set $\tilde{\mathcal{D}}$ without any order constraints between the samples.  

\subsection{Inference} 

During inference, a given test video $V_*$ is partitioned into $m_*$ non-overlapping segments $S_{*j}$, $j \in [1,m_*]$. Feature vectors $\mathbf{f}_{*j}$ are extracted from each segment using $\mathcal{F}(\cdot)$, which are directly passed to the trained detector $\tilde{\mathcal{A}}_{\theta}(\cdot)$ to obtain segment-level anomaly score predictions. Since the eventual goal is frame-level anomaly prediction, all the frames within a segment of the test video are marked as anomalous if the predicted anomaly score for that corresponding segment exceeds a threshold.


\section{Experimental Results}
\label{sec:experiments}

\subsection{Experimental Setup} 

\noindent \textbf{Datasets}: Two large-scale VAD datasets are used to evaluate our approach: UCF-Crime \cite{sultani2018real} and XD-Violence \cite{wu2020not}. \textbf{UCF-Crime} consists of $1610$ ($290$) training (test) videos collected from real-world surveillance camera feeds, totaling 128 hours in length. \textbf{XD-Violence} is a multi-modal VAD dataset that is collected from sports streaming videos, movies, web videos, and surveillance cameras. It consists of $3954$ ($800$) training (test) videos that span around 217 hours. We utilize only the visual modality of the XD-Violence dataset for our experiments. Both these datasets originally contain video-level ground-truth labels for the training set and frame-level labels for the test set. Hence, they are primarily meant for the WS-VAD task. In this work, we ignore the training labels and only use test labels to evaluate our US-VAD model.

\renewcommand{\arraystretch}{1.1}
\begin{table}[t]
\footnotesize
\centering
\begin{center}
\begin{tabulary}{1.0005\linewidth}{CCCCC}
\toprule
\rowcolor{lightgray}
Supervision      & Method                                                                                                       & Features & FNS                                   & AUC(\%)                         \\ \midrule
\multirow{7}{*}{OCC}                                & SVM \cite{sultani2018real}           & I3D  & -    & 50      \\ 
                                                    & Hasan \etal \cite{hasan2016anomaly}         & -        & -    & 50.60   \\   
                                                    & SSV  \cite{sohrab2018subspace}          & -        & -    & 58.50   \\   
                                                    & BODS \cite{wang2019gods}        & I3D  & -    & 68.26   \\   
                                                    & GODS \cite{wang2019gods}           & I3D  & -    & 70.46   \\   
                                                    & SACR  \cite{sun2020scene}        & -        & -    & \textbf{\color{red}{72.70}}   \\   
                                                    & Zaheer \etal.  \cite{zaheer2022generative}  & ResNext  & \xmark   & \textbf{\color{blue}{74.20}}   \\ 
                                                    \hline

\multirow{6}{*}{WS}                                
                                                    & Sultani \etal.$\dagger$ \cite{sultani2018real}     & I3D  & \cmark     & 77.92   \\   
                                                    & Zaheer \etal. \cite{zaheer2022generative}  & ResNext  & \xmark   & 79.84   \\   
                                                    & RTFM    \cite{tian2021weakly}         & I3D  & \cmark    & 84.30   \\   
                                                    & MSL  \cite{li2022self}            & I3D  & \cmark     & \textbf{\color{red}85.30}   \\   
                                                    & S3R \cite{wu2022self}            & I3D  & \cmark     & \textbf{\color{blue}{85.99}}   \\   
                                                    & C2FPL* (Ours)  & I3D  & \xmark   & \textbf{\color{red}{85.5}}      \\ 
                                                    \hline
                                                    \multirow{4}{*}{US}                       & Kim \etal. \cite{kim2021semi}   & ResNext        & -    & 52.00    \\   

                                                     & Zaheer \etal. \cite{zaheer2022generative}  & ResNext  & \xmark    & {71.04}   \\  
                                                     
                                                     & DyAnNet \cite{thakare2023dyannet} & I3D & \cmark & \textbf{\color{red}79.76} \\
                                                    & C2FPL (Ours)  & I3D   & \xmark     & \textbf{\color{blue}{80.65}}  \\
                                                       \bottomrule
\end{tabulary}
\end{center}
\caption{Frame-level AUC performance comparison on UCF-Crime dataset. Wherever available, RGB results are reported. Our unsupervised C2FPL method is compared against both unsupervised and supervised (WS and OCC) methods. The column FNS indicates whether the method uses a fixed number of segments $m$ ($m = 32$ when FNS is true) and `-' indicates this information is not available. The top two results under each supervision setting are shown in blue and red in that order. $\dagger$ indicates that results are reported from \cite{tian2021weakly}, where the method in \cite{sultani2018real} was retrained using I3D features.}
\label{UCF}
\end{table}

\noindent

\noindent \textbf{Evaluation Metric}: We adopt the commonly used frame-level area under the receiver operating characteristic curve (AUC) as the evaluation metric for all our experiments \cite{sultani2018real, zaheer2020claws, zaheer2020cleaning, zaheer2022stabilizing, tian2021weakly, thakare2023rareanom}. Note that the ROC curve is obtained by varying the threshold on the anomaly score during inference and higher AUC values indicate better results. 

\noindent \textbf{Implementation Details}: Each video is partitioned into multiple segments, with each segment containing $r=16$ frames. The well-known I3D~\cite{carreira2017quo} method is used as the pre-trained feature extractor $\mathcal{F}(\cdot)$ to extract RGB features with dimensionality $d = 2048$. Following \cite{takahashi2019data}, we also apply 10-crop augmentation to the I3D features. The CPL generator uses Gaussian Mixture Model (GMM)-based clustering \cite{reynolds2009gaussian} and the threshold $\eta$ is set to $1.0$. The parameter $\beta$ used in the FPL generator is set to $0.2$. The anomaly detector $\tilde{\mathcal{A}}_{\theta}(\cdot)$ is trained using a binary cross-entropy loss function along with $\ell_{2}$ regularization. The detector is trained for $100$ epochs using a stochastic gradient descent optimizer with a learning rate of $0.01$. The batch size $B$ is set to $128$. 

\begin{figure*}[t]
\begin{center}
\includegraphics[width=1\linewidth]{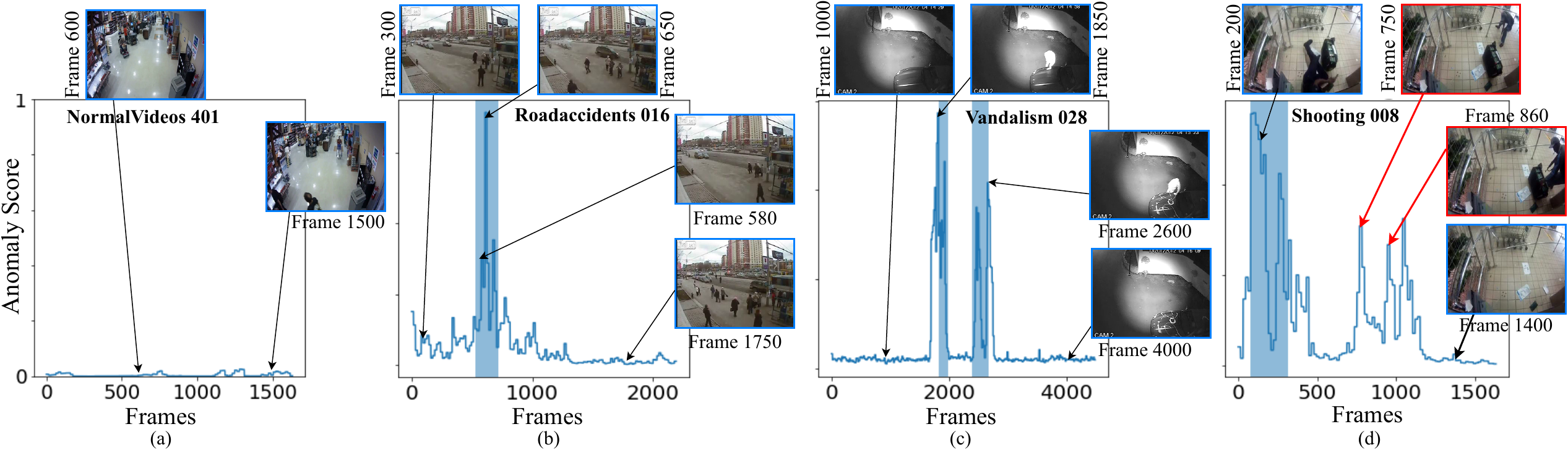} 




\vspace{-1em}
\end{center}
   \caption{Qualitative results of our method on different test videos of the UCF-Crime dataset. The blue color shadow shows the ground truth anomalous frames.}
\label{fig:qual}

\end{figure*}

\renewcommand{\arraystretch}{1.1}
\begin{table}[t]
\footnotesize
\begin{center}
\begin{tabulary}{\linewidth}{CCCCC}
\rowcolor{lightgray}
\toprule
Supervision             & Method        & Features  &  FNS      & AUC(\%)   \\ \midrule
\multirow{5}{*}{OCC} 
                                                    & Hasan \etal \cite{hasan2016anomaly}         & AE        & -    & 50.32   \\  
                                                    & Lu \etal \cite{lu2013abnormal}         & I3D        & -    & 53.56   \\ 
                                                    & BODS \cite{wang2019gods}        & I3D  & -    & \textbf{\color{red}{57.32}}  \\ 
                                                    & GODS \cite{wang2019gods}           & I3D  & -    & \textbf{\color{blue}{61.56}}   \\ 
                                                     
\midrule
\multirow{3}{*}{WS} 
                                                    & S3R \cite{wu2022self}           & I3D       & \cmark    & \textbf{\color{red}{53.52}}   \\ 
                                                    & RTFM$\dagger$ \cite{tian2021weakly}   & I3D        & \cmark    & \textbf{\color{blue}{89.34}}   \\  

                                                     & C2FPL* (Ours)          & I3D       & \xmark   &  \textbf{\color{blue}{90.4}}  \\
                                                    \midrule

\multirow{2}{*}{US} 
& RareAnom \cite{thakare2023rareanom}           & I3D  & \cmark    & \textbf{\color{red}{68.33}}   \\ 

& C2FPL (Ours)  & I3D  & \xmark  & \textbf{\color{blue}{80.09}} \\
\bottomrule

\end{tabulary}
\end{center}
\vspace{-.5em}
\caption{Frame-level AUC performance comparison on XD-Violence dataset. The column FNS indicates whether the method uses a fixed number of segments $m$ ($m = 32$ when FNS is true)  and ``-'' indicates this information is not available. The top two results under each supervision setting are shown in blue and red in that order. $\dagger$ indicates that we
re-compute the AUC of method in \cite{tian2021weakly} using I3D features.}
\label{XD}

\end{table}

\subsection{Comparison with state-of-the-art}

In this section, we provide performance comparison of our proposed unsupervised C2FPL method with recent state-of-the-art (SOTA) supervised and unsupervised VAD methods \cite{sultani2018real,zhang2019temporal,zaheer2022generative,supportVectorMachines,hasan2016anomaly,sohrab2018subspace,wang2019gods}. 

\noindent \textbf{UCF-Crime}. The AUC results on the UCF-Crime dataset are shown in Table \ref{UCF}. Wherever possible, results based on I3D RGB features are reported to ensure a fair comparison. The proposed C2FPL method achieves an AUC performance of $80.65\%$, outperforming the existing US and OCC methods while performing comparably to existing SOTA WS methods. Note that OCC methods assume that the training data contains only normal videos, while we do not make any such assumption. 
Furthermore, our unsupervised C2FPL framework even outperforms some methods in the WS setting \cite{sultani2018real,zhang2019temporal,zaheer2022generative}, thus bridging the gap between unsupervised and supervised approaches. However, compared to the top performing WS method S3R \cite{wu2022self} using the same I3D features, our approach yields $5.34\%$ lower AUC. While this is impressive considering that our method does not require any supervision, it highlights the need for further improvement in the accuracy of the CPL stage. 

\noindent \textbf{XD-Violence}. Our C2FPL framework is also evaluated on XD-Violence dataset and the results are reported in Table \ref{XD}. The proposed method has an AUC of $80.09\%$, which is significantly better than the unsupervised RareAnom \cite{thakare2023rareanom} method. Additionally, our framework achieves good results even in comparison to other OCC and WS methods.

\noindent \textbf{Qualitative Results}: We also provide some qualitative results in Figure \ref{fig:qual}, where anomaly scores predicted by our C2FPL approach are visualized for several videos from the UCF-Crime dataset. It can be observed that the predicted anomaly scores generally correlate well to the anomaly ground truth in many cases, demonstrating the good anomaly detection capability of our approach despite being trained without any supervision. A \textbf{failure case}, shooting008 video (UCF-Crime), is also visualized in Figure \ref{fig:qual}(d). Our detector predicts several frames after the actual shooting event as anomalous. Careful inspection of this video shows a person with a gun entering the scene after the actual event, which our method marks as anomalous, but the ground-truth frame label is normal. Such discrepancies affect the frame-level AUC. 

\setlength{\tabcolsep}{5pt}
\renewcommand{\arraystretch}{1.2}
\begin{table}[t]
\footnotesize
\centering
\begin{center}
\begin{tabulary}{1\linewidth}{C C C C C }
\toprule
\rowcolor{lightgray}
Stage 1 (CPL)         & Stage 2 (FPL)         & Stage 3 (AD)          & Scenario                                                                      & AUC (\%)\\ \midrule
\cmark    & \cmark    & \cmark    & US C2FPL framework                                                                                          & 80.6           \\ \hline
\xmark    & \cmark    & \cmark    & Ground-truth video-level labels (WS)                                                                              & 85.5             \\
\xmark    & \cmark    & \cmark    & Random video-level labels                                                                              & 69.4                \\\hline
\cmark    & \xmark    & \cmark    & CPL pseudo-labels assigned to segments                                                                            & 64.1                  \\
\xmark    & \xmark    & \cmark    & Ground-truth video-level labels assigned to segments                                                             & 72.7                 \\\hline
\xmark    & \xmark    & \cmark    & Random segment-level labels                                                                           & 38.7          \\\hline
\cmark    & \cmark    & \xmark    & (1 – p-value) as anomaly score                                                                           & 57.0               \\
\bottomrule  
\end{tabulary}
\vspace{-1em}
\end{center}
\label{tab:ablation}
\caption{Ablation studies analyzing the impact of each component of the proposed approach on the UCF-Crime dataset.}
\label{tab:ablation}
\end{table}



\subsection{Ablation Study}

Next, we conduct a detailed ablation study to analyze the impact of each component of the proposed C2FPL framework for US-VAD using the UCF-Crime dataset.

\noindent \textbf{Impact of CPL}:
The objective of CPL is to generate coarse video-level labels for all videos in the training dataset. To evaluate the impact of this component, we carry out two experiments and report the results in Table \ref{tab:ablation}. In the first experiment, the CPL stage is removed and the video-level pseudo-labels are assigned randomly. In this case, the performance drops significantly to $69.4\%$ indicating that the coarse pseudo-labels generated by CPL are indeed very useful in guiding the subsequent stages of the proposed system. On the other extreme, we also experimented with using the ground-truth video-level labels instead of the generated coarse pseudo-labels. Note that this setting is equivalent to WS training used widely in the literature. As expected, the performance improves to $85.5\%$, which is almost on par with the best WS method S3R \cite{wu2022self} using the same I3D features (see Table \ref{UCF}). On the XD-Violence dataset, the C2FPL method adapted for the WS setting achieves an AUC of $90.4\%$, which is better than existing WS methods on the same dataset. These results highlight the potential improvement that can be achieved by improving the accuracy of the CPL stage. It also demonstrates the ability of our proposed approach to learn without labels, but at the same time exploit the ground-truth WS labels when they are available.

\noindent \textbf{Impact of FPL}: 
Since the goal of FPL is to obtain segment-level labels, we consider the following three scenarios. Firstly, when C2FPL framework is completely ignored and the segment-level pseudo-labels are assigned randomly, the performance of the trained anomaly detector collapses to a very low AUC of $38.72\%$. This experiment proves that the generated segment-level pseudo-labels are indeed very informative and aid the training of an accurate anomaly detector. Secondly, we ignore only the FPL stage and assign the coarse video-level labels obtained from CPL to all the segments in the corresponding video. There is still a substantial performance drop to $64.1\%$ (from $80.65\%$ when FPL is used). Finally, we again consider the WS setting and assign the ground-truth video-level labels to all the segments in a video. Even in this case, the performance improves only to $72.7\%$ (compared to $85.5\%$ when FPL is used in the WS setting). The last two results clearly prove that the use of FPL reduces segment-level label noise to a large extent, thereby facilitating better training of the anomaly detector. 

\noindent \textbf{Impact of Anomaly Detector}: 
To understand the impact of the segment-level anomaly detector, we excluded the detector and directly used (1–p-value) obtained during the FPL stage as the anomaly score. This results in a significant drop in AUC to $57.0\%$, which indicates that while the C2FPL framework can generate informative pseudo-labels, these labels are still quite noisy and cannot be directly used for frame-level anomaly prediction. The anomaly detector is critical to learn from these noisy pseudo-labels and make more accurate fine-grained predictions.

\begin{table}[h]
\footnotesize
\begin{center}
\begin{tabulary}{0.9\linewidth}{cccccc}
\toprule 
\cellcolor{lightgray}\rotatebox[origin=c]{60}{Method} & \rotatebox[origin=c]{60}{Sultani  \cite{sultani2018real}} & \rotatebox[origin=c]{60}{Zaheer \cite{zaheer2022generative}} & \rotatebox[origin=c]{60}{RTFM \cite{tian2021weakly}} & \rotatebox[origin=c]{60}{S3R \cite{wu2022self}} & \rotatebox[origin=c]{60}{Ours} \\ \hline 
\cellcolor{lightgray}Params & 1.07M & 6.5M & 24.72M & 73.5M & 2.13M \\ \bottomrule
\end{tabulary}
\vspace{-0.5em}

\end{center}
\caption{Number of trainable parameters of the proposed approach in comparison with some existing methods. Our approach achieves good performance with significantly fewer parameters.}
\label{tab:parameters}
\end{table}




\subsection{Parameter Sensitivity Analysis}


\noindent \textbf{Sensitivity to $\eta$}: The sensitivity of the proposed method to the value of $\eta$ is studied on the UCF-Crime dataset. The best results are achieved when $\eta = 1$, which corresponds to having a roughly equal number of videos in the normal and anomaly clusters. When $\eta = 0.5$ or $\eta = 1.5$, the AUC drops to $75.64\%$ and $71.22\%$, respectively. It is important to emphasize that though the number of normal segments in a dataset is usually much larger than the number of anomalous segments, the number of normal and anomalous videos in the available datasets are roughly equal. For example, UCF-Crime has $800$ normal videos and $810$ anomalous videos, while XD-Violence has $2049$ normal videos and $1905$ anomalous videos. Therefore, the choice of $\eta = 1$ is appropriate for these two datasets. In real-world unsupervised settings, the ratio of anomalous to normal videos in a given dataset may not be known in advance because there are no labels. When $\eta$ is mis-specified, there is some performance degradation, which is a limitation of the proposed C2FPL approach.

\begin{figure}[t]

\begin{center}
    \includegraphics[width=1\linewidth]{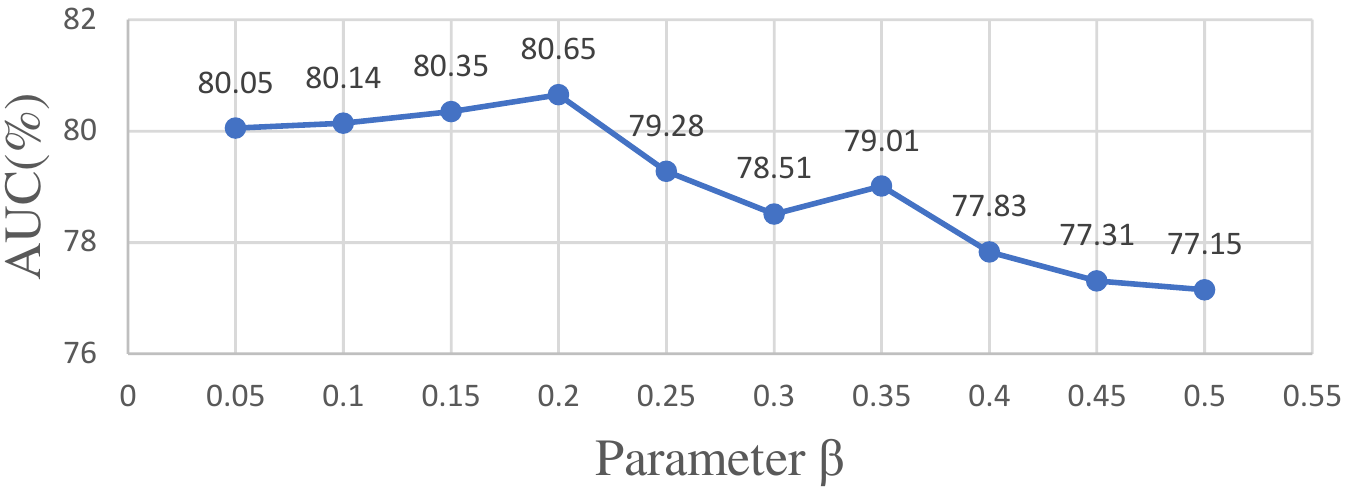}    
\end{center}
\vspace{-1em}

\caption{Sensitivity of C2FPL framework to parameter $\beta$.}
\label{fig:windowsize}

\end{figure}

\noindent \textbf{Sensitivity to $\beta$}: In the FPL stage, a window of size $\lceil\beta m_i\rceil$ is used to \textit{loosely} incorporate the temporal contiguity constraint. In the earlier experiments, $\beta$ was set to $0.2$ ($20\%$ of the video length). However, in practice, the number of anomalous segments in a video may vary widely and would not be known in advance. The sensitivity of the proposed method to the value of $\beta$ is shown in Figure \ref{fig:windowsize}. These results indicate that our method is quite robust to changes in $\beta$.   

\noindent \textbf{Fixed number of segments}: We hypothesize that not compressing the videos at test/train time, as commonly done in the existing literature \cite{sultani2018real,tian2021featuremagnitude}, is beneficial for the overall anomaly detection performance. To validate this hypothesis, we experiment with compressing each video into to a fixed number of segments $m=32$ before applying the proposed method. With such compression, the performance of our method drops to $77.70\%$ and $78.08\%$ for UCF-Crime and XD-Violence datasets, respectively. This justifies our choice of not using any compression.

\subsection{Computational Complexity Analysis}


Apart from feature extraction, the proposed C2FPL training method requires a few invocations of the GMM clustering subroutine, a single round of Gaussian distribution fitting, and training of the segment-level anomaly detector $\tilde{\mathcal{A}}_{\theta}$. Since GMM clustering is performed at the video level on 2D data, the computational cost of the two-stage pseudo-label generator is insignificant ($0.6$ seconds) compared to that of the anomaly detector training ($60$ seconds per epoch). As seen in Figure \ref{fig:arch}, the architecture of $\tilde{\mathcal{A}}_{\theta}$ is fairly simple with only $2.13$M parameters, which is significantly lower than all SOTA methods except Sultani \etal \cite{sultani2018real}, as shown in Table \ref{tab:parameters}. It may be noted that, despite having fewer parameters, the WS variant of our approach outperforms almost all the other methods on both datasets (Table \ref{UCF} \& \ref{XD}). The only exception is S3R, which has $0.5\%$ higher AUC compared to our approach, while having over $71$M extra parameters than our method. During inference, our method achieves 70 frames per second (fps) on NVIDIA RTX A6000 which is almost double the rate of real-time applications. This indicates that our system can achieve good real-time detection in real-world scenarios.









\section{Conclusion}
Unsupervised video anomaly detection (US-VAD) methods are highly useful in real-world applications as a complete system can be trained without any annotation or human intervention. In this work, we propose a US-VAD approach based on a two-stage pseudo-label generator that facilitates the training of a segment-level anomaly detector. Extensive experiments conducted on two large-scale datasets, XD-Violence and UCF-Crime, demonstrate that the proposed approach can successfully reduce the gap between unsupervised and supervised approaches.


{\small
\bibliographystyle{ieee_fullname}
\bibliography{WACV_Template}
}

\newpage
\appendix
\appendixpage


\section{Self Attention}
\begin{figure}[t]
\begin{center}
    \includegraphics[width=1\linewidth]{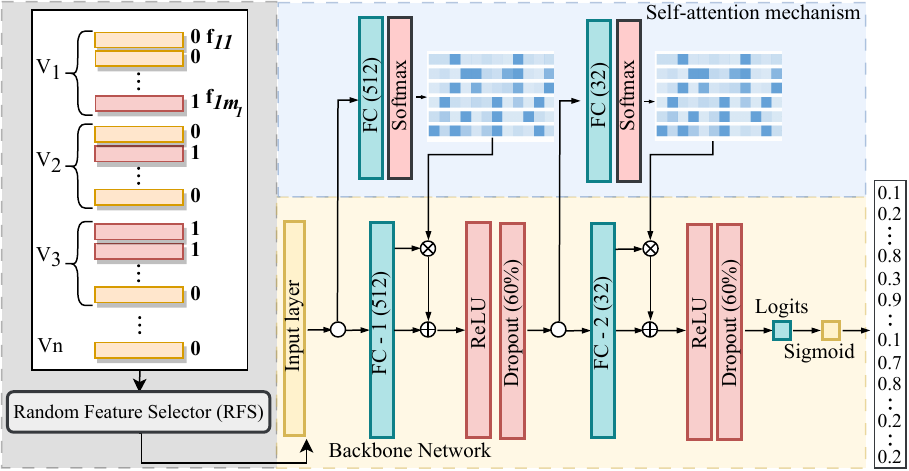}
\end{center}
   \caption{
 Detailed architecture of our proposed learning network: The training batch containing pseudo-labeled feature vectors is the input to the FC backbone network (lower). In addition to the backbone
network, we add two self-attention layers (upper).}
\label{fig:detector}
\end{figure}

Figure \ref{fig:detector} shows the detailed architecture of our proposed C2FPL network. The FC layers described in manuscript: Section 3.3 have 512 and 32 neurons where each is followed by a ReLU activation function and a dropout layer with a dropout rate of 0.6. In addition, we add two self-attention layers. In this section, we will discuss the choice design as well as the aim of using this layer.

\begin{figure}[t]
\centering

\begin{subfigure}{0.9\linewidth}
  \centering
  \includegraphics[width=\linewidth]{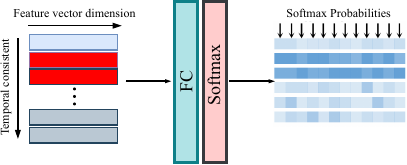}
  \caption{Batch Dimension (BD) SA}
  \label{fig:subfig4}
\end{subfigure}
\hfill
\begin{subfigure}{0.9\linewidth}
  \centering
  \includegraphics[width=\linewidth]{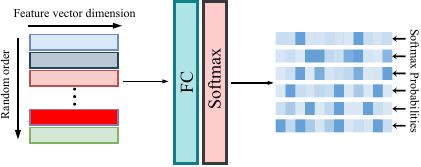}
  \caption{Features Dimension (FD) SA}
  \label{fig:subfig3}
\end{subfigure}

\caption{Visualization of the two types of self-attention: (a) Batch Dimension (BD): Softmax probabilities
    are calculated along the Batch dimension (temporal axis). (b) Features Dimension (FD): Softmax probabilities
    are calculated along the feature vector dimension.}
\label{fig:SA_type}
\end{figure}

The aim of self-attention (SA) in our proposed C2FPL framework is to highlight parts of feature vectors critical in detecting anomalies. 
Our configuration applies self-attention over each feature vector (feature dimension) independently without requiring temporal order. This is unlike a compareable existing architecture by Zaheer \etal \cite{zaheer2020claws} where the Normalcy Suppression Module (NSM) aims to learn attention based on the temporally consistent feature vectors in the input batch (Figure \ref{fig:SA_type}(a)) and the attention is calculated along the batch dimension (temporal axis).



To study this in details, we define several possible configurations of the self-attention used in our C2FPL and report their performances in this section. Through thorough analysis, we verify the effectiveness of our design choices within the framework.

\begin{table}[b]
\small
\begin{center}
\begin{tabular}{c c c}
\toprule
\rowcolor{lightgray} Framework         & SA configuration & AUC (\%) \\ \midrule
\multirow{2}{*}{$\text{CPL} \rightarrow \text{FPL} \rightarrow \text{AD}$} & Multiplicative                    & 63.5     \\ \cline{2-3} 
                  & Residual  (Ours)               & 80.6     \\ \hline
\end{tabular}
\end{center}
\caption{Area under the curve (AUC) comparison of two SA configurations
configurations on the UCF-Crime dataset. (The framework configuration is the same as shown in manuscript: Table 3). }
\label{SA_config}
\end{table}

\begin{table}[t]
\small
\begin{center}
\begin{tabular}{ccc}

\toprule
\rowcolor{lightgray}Framework         & SA Dimension & AUC (\%) \\ \midrule
\multirow{2}{*}{$\text{CPL} \rightarrow \text{FPL} \rightarrow \text{AD}$} & Batch Dimension                    & 76.5     \\ \cline{2-3} 
                  & Feature Dimension  (Ours)               & 80.6     \\ \hline
\end{tabular}
\end{center}
\caption{Area under the curve (AUC) comparison of two SA types on UCF-Crime dataset. (The framework is the same as shown in manuscript: Table 3). }
\label{SA_UCF}
\end{table}

\subsection{Residual vs Multiplicative Self-Attention (SA)}
\begin{figure}[t]
\begin{center}
    \includegraphics[width=0.9\linewidth]{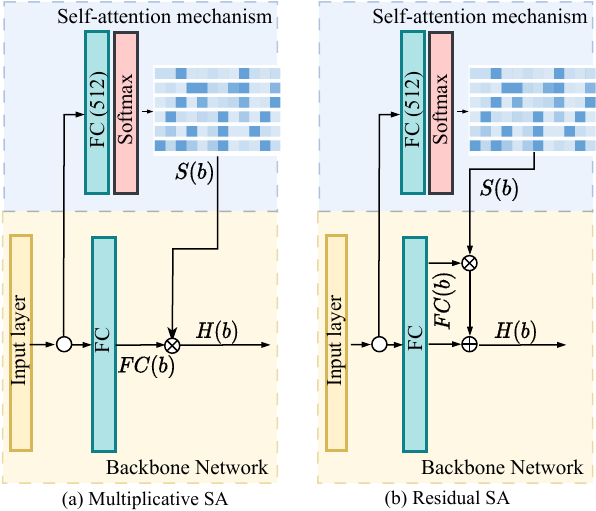}
\end{center}
   \caption{
Visualization of the two self-attention configurations including (a) Multiplicative SA and our proposed (b) Residual SA.}
\label{fig:SA_config}
\end{figure}

Zaheer \etal \cite{zaheer2020claws}, in CLAWS Net, formulate the problem of self-attention in terms of suppressing certain features which are achieved by multiplicative attention. To provide a comparison, we discuss two different SA configurations as shown in Figure \ref{fig:SA_config}. First, following Zaheer \etal \cite{zaheer2020claws}, given an input batch $b$ we calculate the output $H(b)$ by performing an element-wise multiplication $\otimes$ between SA output $S(b)$ and backbone output $FC(b)$ as:
$$
H(b)=S(b) \otimes FC(B)
$$

Although such multiplication has been helpful in CLAWS Net, generally it has been shown to have the unfavorable result of dissipating model representations \cite{hu2018squeeze, wang2017residual}. It's because attention generates probabilities that, when multiplied by the features directly, can drastically lower the values.

In our framework, we utilize residual SA in which attention-applied features are added back to the original features. Therefore, The  output $H(b)$ is calculated as:
$$
H(b)= (FC(b) \otimes S(b)) \oplus FC(b)
$$
where $\oplus$ is an addition operation. 

Table \ref{SA_config} shows the performance difference between multiplication and residual attention approaches. We can observe that the use of multiplication negatively affects our model's AUC performance (63.5\%). We attribute this to the suppression nature of multiplication \cite{hu2018squeeze, wang2017residual}. 
The specifically designed NSM of CLAWS Net \cite{zaheer2020claws} aims to dissipate normal portions of the temporally consistent input batches that help the backbone network produce low anomaly scores. 
However, the nature of our training is not suitable for this formulation. Therefore, using residual attention, which only highlights individual parts of each feature vector in a given batch, the performance of our model increases to 80.6\% on the UCF-crime dataset.

\begin{figure*}[t]
\begin{center}
    \includegraphics[width=1\linewidth]{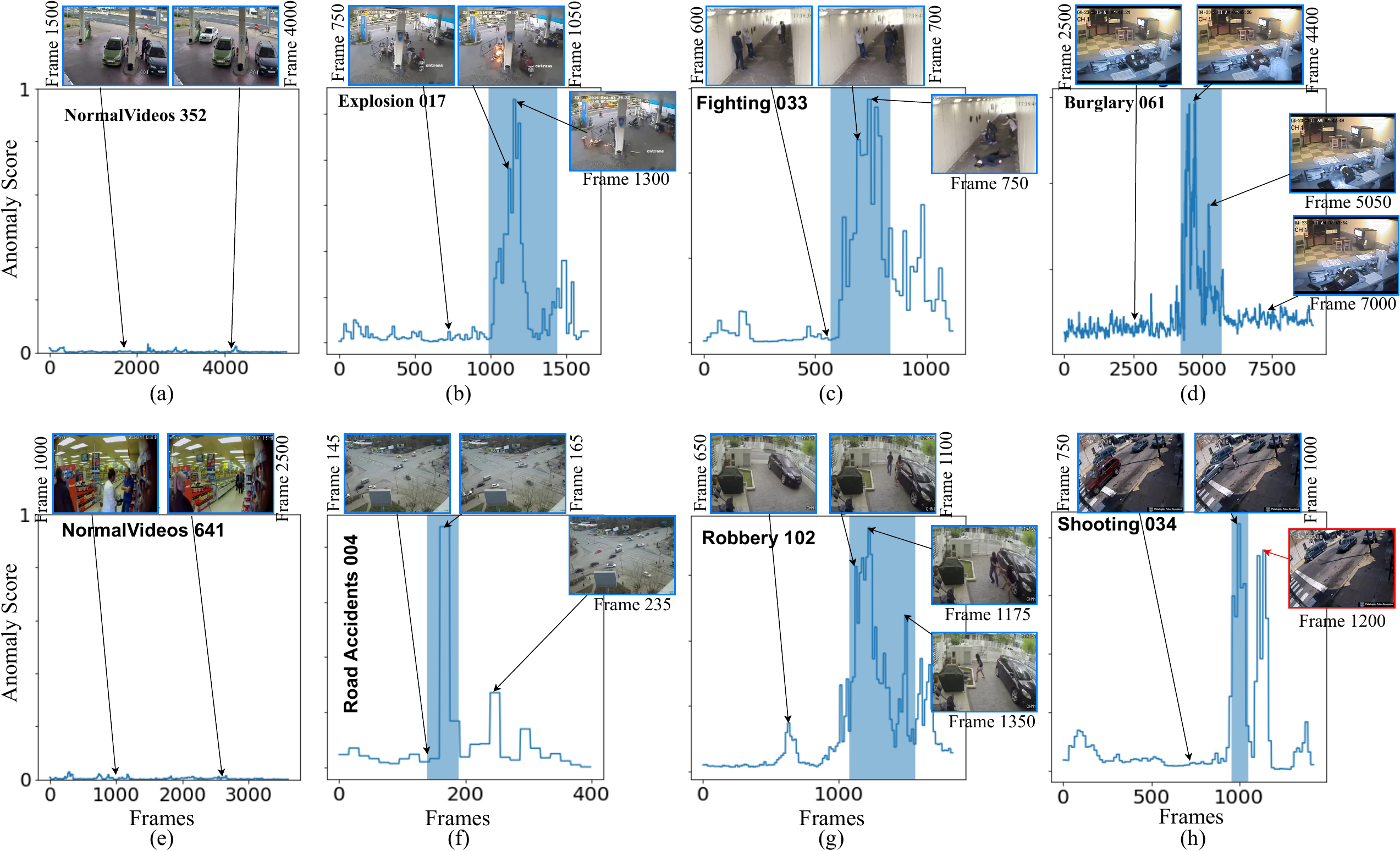}
\end{center}
   \caption{
Anomaly scores of the proposed C2FPL framework on different videos from the UCF-Crime Dataset. }
\label{fig:qual1}

\end{figure*}

\subsection{Types of self-attention}
In conjunction with Zaheer \etal \cite{zaheer2020claws}, We discuss two different types of self-attentions depending on the dimensions along which Softmax probabilities are computed in an element-wise fashion.

\textbf{Softmax probabilities over the batch dimension (BD)}. 
As mentioned, Zaheer \etal \cite{zaheer2020claws} calculates the probabilities temporally to make use of the temporal information preserved within a batch (Figure \ref{fig:SA_type} (a)). However, we have argued and demonstrated in our presented C2FPL framework that preserving temporal information is not necessary for improved anomaly detection performance. Therefore, using temporal attention along the batch dimension may not be as effective in our framework as it has been proven in CLAWS Net by Zaheer \etal \cite{zaheer2020claws}. Nevertheless, we utilize their proposed self-attention and compare it with our design of self-attention.

\textbf{Softmax probabilities over the feature dimension (FD)}. This self-attention over feature dimension (FD) is the configuration used in our C2FPL framework, as explained in manuscript: Section 3.3 (lines 494-500).
Since we assume no temporal consistency among batches, the probabilities are computed over the feature dimension (Figure \ref{fig:SA_type} (b)).

Table \ref{SA_UCF} summarizes the frame-level AUC performance of the two types. It can be seen that the FD type (ours) outperforms the BD type attention by a margin of 4.1\%. 
This verifies the importance of using self-attention along the feature vector dimension, achieving significant performance
gains.

\section{Qualitative Results} We also provide additional qualitative results in Figure \ref{fig:qual1}, where anomaly scores predicted by our C2FPL approach are visualized for other classes of anomalous videos from the UCF-Crime dataset. In some cases, the anomalous frames in certain videos might exceed the annotated ones because the annotations only cover a portion of the event. For instance, the abnormal event in the RoadAccidents004 video begins at about frame 145 and lasts significantly longer than the annotated window, which only shows the accident impact event.

An additional \textbf{failure case}, shooting034 video (UCF-Crime), is also visualized in Figure \ref{fig:qual1}(h). Our proposed model correctly predicts the ground-truth anomalous window. However, later frames (1200) of the video show one of the occupants involved in the shooting quickly entering his car before speeding off, which our detector marks as an anomalous event while that event is annotated as a normal event.

\section{Convergence Analysis}
\begin{figure}[t]
\begin{center}
    \includegraphics[width=1\linewidth]{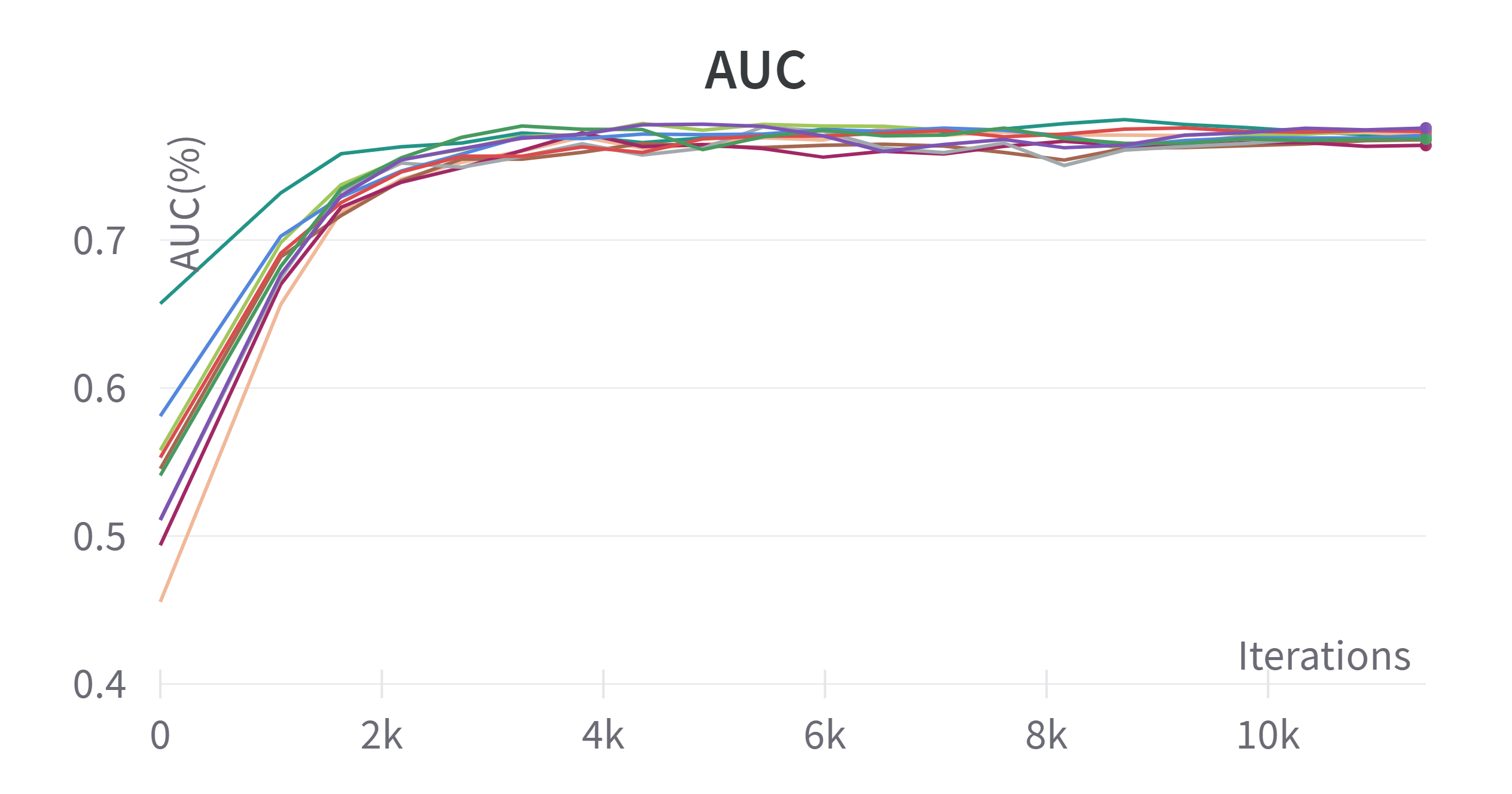}
\end{center}
   \caption{
 Convergence of our proposed model using multiple random seed experiments. }
\label{fig:conv}

\end{figure}

As our approach is an unsupervised anomaly detection method, we empirically analyze its convergence using 10 random seed runs as shown in Figure \ref{fig:conv}. For all experiments, our C2FPL model attains an average AUC of 80.14\% $\pm$ 0.31\%. This demonstrate that our proposed framework not only achieves excellent anomaly detection but also demonstrates good convergence.

\end{document}